\documentclass{article}

\usepackage[preprint]{neurips_2020}

\setcitestyle{citesep={,}}

\usepackage[utf8]{inputenc} % allow utf-8 input
\usepackage[T1]{fontenc}    % use 8-bit T1 fonts
\usepackage{subcaption}
\usepackage{url}            % simple URL typesetting
\usepackage{booktabs}       % professional-quality tables
\usepackage{amsfonts}       % blackboard math symbols
\usepackage{nicefrac}       % compact symbols for 1/2, etc.
\usepackage{graphicx}
\usepackage{microtype}      % microtypography
\usepackage{float}
\usepackage{wrapfig}
\usepackage{tabularx}
\usepackage{textcomp}
\usepackage{gensymb}
\usepackage{amsmath}

\title{Benchmarking Deep Inverse Models over time, and the Neural-Adjoint method}

\author{
  Simiao Ren\\\\
  Dept. of Electrical and Computer Engineering\\
  Duke University\\
  Durham, NC 27705 \\
  \texttt{simiao.ren@duke.edu} \\
  \And
   Willie J. Padilla \\\\
  Dept. of Electrical and Computer Engineering\\
  Duke University\\
  Durham, NC 27705 \\
  \texttt{willie.padilla@duke.edu} \\
  \And
  Jordan Malof \\
  Dept. of Electrical and Computer Engineering\\
  Duke University\\
  Durham, NC 27705 \\
  \texttt{jordan.malof@duke.edu} \\
}

\begin{document}

\maketitle
\setcitestyle{square}
\begin{abstract}
  We consider the task of solving generic inverse problems, where one wishes to determine the hidden parameters of a natural system that will give rise to a particular set of measurements.  Recently many new approaches based upon deep learning have arisen, generating promising results.  We conceptualize these models as different schemes for efficiently, but randomly, exploring the space of possible inverse solutions.  As a result, the accuracy of each approach should be evaluated as a function of time rather than a single estimated solution, as is often done now.  Using this metric, we compare several state-of-the-art inverse modeling approaches on four benchmark tasks: two existing tasks, a new 2-dimensional sinusoid task, and a challenging modern task of meta-material design.  Finally, inspired by our conception of the inverse problem, we explore a simple solution that uses a deep neural network as a surrogate (i.e., approximation) for the forward model, and then uses backpropagation with respect to the model input to search for good inverse solutions.  Variations of this approach - which we term the neural adjoint (NA) - have been explored recently on specific problems, and here we evaluate it comprehensively on our benchmark.  We find that the addition of a simple novel loss term - which we term the boundary loss - dramatically improves the NA's performance, and it consequentially achieves the best (or nearly best) performance in all of our benchmark scenarios. 
\end{abstract}

\section{Introduction}
In this work we consider the task of solving generic inverse problems. An inverse problem is characterized by a forward problem that models, for example,  a real-world measurement process or an auxiliary prediction task. The forward problem can be written as
\begin{equation}
    y = f(x)
\end{equation}
where $y$ is the measurable data, $f$ is a (non-)linear forward operator that models the measurement process, and $x$ is an unobserved signal of interest. Given $f$, solving the inverse problem is then a matter of finding an inverse model $x = f^{-1} (y)$. However, if the problem is ill-posed (e.g., non-existence, or non-uniqueness, of solutions), finding $f^{-1}$ is a non-trivial task. Specific inverse problems can be solved using apriori knowledge about $f$, (e.g., sparsity in some basis, such as compressed sensing), however, we consider the task of solving generic inverse problems, where no such solutions are known.  

Recently many new approaches based upon deep learning have arisen, generating impressive results.  These methods typically require a dataset of sample pairs $\{x_n,y_n\}_{n=1}^N$ from $f$, from which a deep neural network model can be trained to approximate the inverse model, $ \hat{f}^{-1}$.  Some recent examples include models based on normalizing flows (e.g., invertible neural networks \citep{ardizzone2018analyzing,kruse2019benchmarking}), variational auto-encoders \citep{kingma2013auto}, tandem architectures \citep{liu2018training,jordan1992forward}.   

\subsection{Modern inverse models as stochastic search}
Despite the apparent variety of recent approaches, most of these inverse models can be written in the form $\hat{x} = \hat{f}^{-1}(y,z)$, where $z$ is randomly drawn from some probability distribution $Z$ (e.g., Gaussian). Although the interpretation of $z$ varies across these models, they all share the property that the $\hat{x}$ returned by the model will vary depending upon the value of $z$.  Furthermore, since it is usually trivial and fast to evaluate the accuracy of a candidate inverse solution using the forward model, $f$ (e.g., a simulator), one can search for more accurate inverse solutions by sampling multiple values of $z$, each yielding a different inverse solution.  Each solution can then be validated using $f$, and the best solution among all candidates can be retained. Therefore, each modern inverse model can be viewed as a means of efficiently, but nonetheless stochastically, searching through x-space for good  solutions.   

From this perspective, the performance of each inverse model depends upon the number of $z$ samples that are considered, denoted $T$.  For example, one model may perform best when $T=1$, while another model performs best as $T$ grows.  Our experiments here show that this is indeed the case, and model performance (relative to others) is highly dependent upon $T$. Typically however the performance, $r$, of an inverse models is judged by estimating its expected ``re-simulation'' error \citep{kruse2019benchmarking} over the data and latent variable distributions, denoted $D$ and $Z$ respectively.  Mathematically, we have 
\begin{equation}
    r = E_{(x,y)\sim D, z \sim Z}[\mathcal{L}(\hat{y}(z),y)] \label{eq:r}
\end{equation}
where $\hat{y}(z) = f(\hat{f}^{-1}(y,z))$ is the "re-simulated" value of $y$ produced by passing $\hat{x}$ (an estimate) through the forward model, and $\mathcal{L}$ is the user-chosen loss function (e.g., L2 loss).  The metric $r$ effectively measures error under the assumption we always utilize one sample of $z$ (given a target $y$). Here we propose an alternative metric that quantifies the expected \textit{minimum}  error if we draw a sequence of $z$ values of length $T$, denoted $Z_{T}$. Formally, this is given by  
\begin{equation}
    r_T = E_{(x,y)\sim D, Z_T \sim \Omega}\Big[ \min_{z\in Z_T}[\mathcal{L}(\hat{y}(z),y)] \Big] \label{eq:rt}
\end{equation}
where $Z_T$ is a sequence of length $T$ drawn from a distribution $\Omega$. This measure characterizes the expected loss of an inverse model \textit{as a function of} the number of samples of $z$ we can consider for each target $y$.  In this work we conduct a benchmark study of four tasks with $r_T$, and we find that the performance of modern inverse models depends strongly on $T$, revealing the limitation of existing metrics, and revealing useful insights about the way in which each model stochastically searches $x$-space.  In particular, we present analysis suggesting that modern inverse models suffer from one or both of the following limitations in their search process: (i) they don't fully explore $x$-space, missing some solutions; or (ii) they do not precisely localize the optimal solutions, introducing error.     

\subsection{The neural-adjoint method}
Inspired by our conception of the inverse problem, we explore a simple solution where the main idea is to train a neural network to approximate $f$ and then, starting from different random locations in $x$-space, use $\partial \hat{f} / \partial x$ to descend towards locally optimal $x$ values. Variations of this approach have recently been employed on a few specific problems \citep{gomez2018automatic, peurifoy2018nanophotonic}, however, here we evaluate its competitiveness against other modern approaches on several tasks.  We also add a novel simple term to its loss function - which we term the boundary loss - that dramatically improves its performance.   We call the resulting method the Neural Adjoint (NA), due to its resemblance to the classical Adjoint method for inverse design \cite{bendsoe1988generating,herskovits1995advances}.  Surprisingly, the relatively simple NA approach almost always yields the lowest error among all models, tasks and $T$-values considered in our benchmarks.  Our analysis suggests that, in contrast to other models, NA fully explores the $x$-space, and also accurately localizes inverse solutions.  NA achieves this advantage at the cost of significantly higher computation time, which as we discuss, may disqualify it from some time-sensitive applications.   

In summary, the three primary contributions of this work are as follows:
\begin{enumerate}
  \item \textit{A comprehensive benchmark comparison using $r_{T}$.}  We compare five modern inverse models on four benchmark tasks.  The results reveal the performance of modern models under many different conditions, and we find that their accuracy depends strongly on $T$.   
  \item \textit{A new modern benchmark task, and a general method to replicate it}. We introduce a contemporary and challenging inverse problem for meta-material design. Normally, it would be difficult for others to replicate our studies because requires sophisticated electromagnetic simulations.  However, we introduce a strategy for creating simple, fast, and sharable \textit{approximate} simulators for complex problems, permitting easy replication.  
  \item \textit{The neural-adjoint (NA) method.} The NA nearly always outperforms all other models we consider in our benchmark.  Furthermore, our analysis provides insights about the limitations of existing models, and why NA is effective.    
\end{enumerate}
We release code for all inverse models, as well as (fast) simulation software for each benchmark problem, so that other researchers can easily repeat our experiments. \footnote{\url{https://github.com/BensonRen/BDIMNNA}}

\section{Related Work}
\textbf{Modern deep inverse models}. Given some samples from a forward model, learning the inverse mapping is difficult even for trivial tasks because of one-to-many mappings, where several input values (e.g., designs) all give rise to the same (or similar) forward model output \citep{maass2019deep}.  This causes problems with many optimizers and loss functions because they assume a unimodal output.  For example, using gradient descent with mean-squared error causes the model to produce solutions that are an average of all individual solutions, which is usually not a valid solution.  To address this inconsistent gradient information, cyclic consistent loss or Tandem models \citep{liu2018training,song2018learning,zhu2017unpaired,pilozzi2018machine} avoid this dilemma by connecting a forward model to the backward model, thereby effectively backpropagating using only one solution, even if multiple solutions exist.  An alternative approach is to model the conditional posterior, $p(x|y)$, directly using variational methods \citep{ma2019probabilistic,kiarashinejad2020deep}. Variational Auto-Encoders (VAEs) \citep{kingma2013auto} consist of an encoder and decoder, and model the joint distribution of hidden and measurement states, to normal distributions z, and decode inverse solutions from samples. By minimizing the evidence lower bound, it trades between reconstruction accuracy and transformed joint distribution closeness to a normal distribution. Earlier work on Mixture density networks (MDNs) \citep{bishop1994mixture}  directly model the conditional distribution using a mixture of gaussian distributions. The parameters of the gaussians are predicted by a feedforward neural network. With recent advance in the normalizing flow community \citep{tabak2010density,germain2015made,dinh2014nice,ardizzone2018analyzing} first applied a state-of-the-art invertible neural network to the inverse problem. Utilizing various invertible network-based architectures, Kruse \citep{kruse2019benchmarking} benchmarked them on two simple inverse problems. It was found that conditional invertible networks (cINN), and invertible networks (INN) trained by maximum likelihood, had the best performance.  Many of these models were recently benchmarked on two inverse problems in \citep{kruse2019benchmarking}: VAE, INN,cINN, and MDN.  We reproduce their results here, but we add a tandem model and the NA model.  We also compare all models on two additional benchmark tasks introduced here (i.e., four total task).  

\textbf{Inverse model performance metrics}. Although the architecture varies across different studies, the performance metric used in each is largely identical. Nearly all the studies on inverse regression problems uses either Mean Squared Error or Root Mean Squared Error \textit{with only one evaluation}, \citep{ardizzone2018analyzing,kruse2019benchmarking,liu2018training,maass2019deep,pilozzi2018machine,ma2019probabilistic,kiarashinejad2020deep,peurifoy2018nanophotonic} despite the stochastic nature of some approaches, which can produce different solutions for the same target. Posterior matching is less of a focus in this paper and the Maximum Mean Discrepancy (MMD) score is appended in the supplement.

\textbf{Adjoint-based methods}.  The adjoint method is a popular approach in control theory and engineering design that relies upon finding an analytical gradient of the forward model with respect to the controllable variables, and then using this gradient to identify locally optimal inverse solutions.  The NA method here also utilizes gradients of the forward model to identify locally optimal inverse solutions, however, by using a neural network to approximate the forward model (and its gradients) there is no need to derive an analytic expression.  Variants of this strategy have also recently been employed by \citep{peurifoy2018nanophotonic} for meta-material design (our inspiration), and \cite{gomez2018automatic} in molecule design.  We primarily build upon their work by (i) distilling and describing the essential elements of this approach; (ii) introducing the boundary loss, and conducting comprehensive experiments that show it substantially improves the accuracy and reliability of this approach; and (iii) conducting a comprehensive comparison of the resulting approach (the NA method) against other modern models.

\section{The Neural-Adjoint Method} \label{sec:neuraladjoing}
\begin{figure}[H]
  \centering
  \includegraphics[width=13cm]{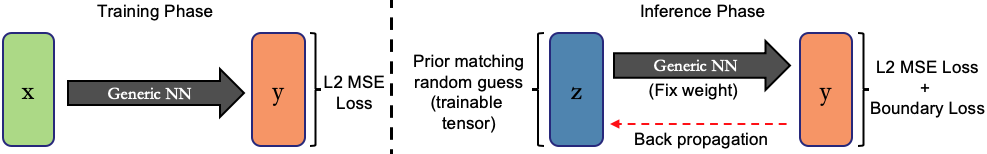}
  \caption{Architecture of Neural Adjoint method}
  \label{fig:BP}
\end{figure}

The NA method can be divided into two steps: (i) Training a neural network approximation of $f$, and (ii) inference of $\hat{x}$. Step (i) is conventional and involves training a generic neural network on a dataset of input/output pairs from the simulator, denoted $D$, resulting in $\hat{f}$, an approximation of the forward model.  This is illustrated in the left inset of Fig \ref{fig:BP}.  In step (ii), our goal is to use $\partial \hat{f} / \partial x$ to help us gradually adjust $x$ so that we achieve a desired output of the forward model, $y$.  This is similar to many classical inverse modeling approaches, such as the popular Adjoint method \citep{bendsoe1988generating,herskovits1995advances}.  For many practical inverse problems, however, obtaining $\partial \hat{f} / \partial x$ requires significant expertise and/or effort, making these approaches challenging.  Crucially, $\hat{f}$ from step (i) provides us with a closed-form differentiable expression for the simulator, from which it is trivial to compute $\partial \hat{f} / \partial x$, and furthermore, we can use modern deep learning software packages to efficiently estimate gradients, given a loss function $\mathcal{L}$. 

More formally, let $y$ be our target output, and let $\hat{x}^{i}$ be our current estimate of the solution, where $i$ indexes each solution we obtain in an iterative gradient-based estimation procedure.  Then we compute $\hat{x}^{i+1}$ with  
\begin{equation}    
\hat{x}^{i+1} = \hat{x}^{i} - \alpha  \left. \frac{\partial \mathcal{L} (\hat{f}(\hat{x}^{i}),y)}{\partial x} \right\rvert_{x=\hat{x}^{i}} \label{eq:nagrad} 
\end{equation}
where $\alpha$ is the learning rate, which can be made adaptive using conventional approaches like Adam \citep{kingma2014adam}.  Notice that the parameters of the neural network are \textit{fixed}, and we are only adjusting the input to the network, treating them like model parameters. Our initial solution, $\hat{x}^{0}$ is drawn from some distribution $\Gamma$.  Given some desired $y$, NA iteratively adjusts its estimated solution (beginning with $\hat{x}_{0}$) until convergence (e.g., $\mathcal{L}$ no longer reduces).  This entire process acts as the inverse model for the process, $\hat{f}^{-1}(y,z)$, where $z=\hat{x}^{0}\sim\Gamma$.  This is illustrated in the right inset of Fig \ref{fig:BP}.  Similar to other approaches, we can draw a sequence of $z$ values and obtain an estimated solution for each one.  

And as we show in our experiments in Section \ref{sec:bestmodels}, the NA method yields highly accurate solutions compared to other models, however, at the cost of relatively high computation time.   One challenge with this approach is that many initializations either (i) do not finish converging, or (ii) converge to a poor minima. To mitigate this problem, we always extract a thousand solutions, and use the NA's built-in forward model to internally rank-order the solutions and return only the top "T" solutions to be evaluated by the true simulator. As we discuss in Section \ref{sec:bestmodels}, this process only marginally increases the inference time of NA (and all inverse methods we consider) because of efficient parallel processing on GPUs.  However, because the NA uses an iterative gradient descent procedure, it is still computationally expensive compared to other methods.

\subsection{Obtaining good results: the boundary loss} \label{sec:naimproved}
Another challenge with NA is that (unless restricted) it frequently converges to solutions that are outside of the training data sampling domain.  As we show in the supplement, this seems to occur because $\hat{f}$ becomes highly inaccurate outside of the training data domain, and (erroneously) predicts that $x$-values in this space will produce the desired $y$ (or a close approximation).  As a consequence, these inverse solutions are generally inaccurate, resulting in high error when evaluated with the true simulator. To discourage this behavior we add a simple ``boundary loss'' term that encourages NA to identify solutions that are within the training data domain, where $\hat{f}$ is accurate.  This loss term, denoted $\mathcal{L}_{bnd}$, is given by 
\begin{align}
    \label{eq:1}
    \mathcal{L}_{bnd} &= ReLU(|\hat{x} - \mu_x| - \frac{1}{2}R_x)
\end{align}
where $\mu_x$ is the mean of the training data, $R_x$ is its range (for unbounded distributions of $x$, we define the range to be the interval of 95\% probability), and ReLU is the conventional neural network activation function.  This loss is only added during the inference of inverse solutions. As we show in the supplement, without $\mathcal{L}_{bnd}$ added, the performance of NA decreases substantially. In the supplement we also visualize the NA method, and without, $\mathcal{L}_{bnd}$ on a simple 1-dimensional task, illustrating its effects.        

\textbf{Some limitations}.  Although effective, the form of the boundary loss in eq. \ref{eq:1} assumes that the training data domain is well approximated by a hyper-cube of the form $|\hat{x} - \mu_x| - 0.5R_x$.  If this is not true (e.g., the domain is non-convex) then $\mathcal{L}_{bnd}$ may become less effective.  Furthermore, the form of $\mathcal{L}_{bnd}$ implicitly assumes $\hat{f}$ is uniformly accurate within the training domain, and drops equally in all directions outside of it. However, the accuracy of $\hat{f}$ will not generally meet these assumptions, and it is unclear how the loss in $\mathcal{L}_{bnd}$ would vary as a function of the uncertainty of $\hat{f}$.  

\textbf{Relationship to other methods}.  \textit{Trust region optimization (TRO)} \citep{kurutach2018model}. The boundary loss and TRO both identify regions where $\hat{f}$ is accurately approximating $f$, called a "trust region", and use this information to guide the search for solutions.  However, the NA uses a static trust region that encompasses the whole training data domain, while TRO estimates local trust regions during the solution search process.   \textit{Bayesian Optimization (BO)} \cite{mockus1978application, snoek2012practical}. In BO, $\mathcal{L}_{bnd}$ can be interpreted as a prior on the credible interval of the surrogate function ($\hat{f}$ in our case) that is uniformly valued within the training data domain, and then grows outside of it.  In BO this prior might cause the acquisition function to sample $f$ at these locations and update $\hat{f}$, however in contrast we use it to discourage the acquisition function (gradient descent in our case) from seeking solutions in these regions. \textit{Adaptive sampling} \citep{brookes2019conditioning}.  In adaptive sampling $\hat{f}$ is progressively updated using samples at locations where it is estimated to be inaccurate.  In contrast,  $\mathcal{L}_{bnd}$ essentially assumes the model is equally accurate throughout the training data domain and, similar to BO, $\hat{f}$ is not updated with samples from $f$.  

\section{Benchmark deep inverse models} \label{sec:inversemodels}
In this section we briefly describe the inverse models that we employ in our benchmark experiments. We focus on the motivation and relevant properties of each model, however, more detail for each model can be found in the supplement and in referenced supplied for each method. 

\textbf{Mixture Density Networks (MDN)} \citep{bishop1994mixture}.  MDNs model the conditional distribution $p(x|y)$ as a mixture of Gaussians, parameterized by $\mu_i, \Sigma_i$ and $p_i$ (mixing proportion). A neural network is trained to predict the \textit{parameters} of the mixture, given a $y$ value, using the following loss:
\begin{equation}
    Loss = - \log(  \sum_i p_i *|\Sigma_i^{-1}|^{\frac{1}{2}} * \exp(-\frac{1}{2} (\mu_i - x)^{T} \Sigma_i^{-1} (\mu_i - x)))
\end{equation}
The number of Gaussians is a hyper-parameter. Once the parameters are predicted for a given $y$, then $\hat{x}$ are inferred by randomly sampling the mixture distribution, and therefore each sample represents a different $z$ value in the stochastic search process.    

\textbf{Conditional Variational Auto-Encoder (cVAE)}. \citep{ma2019probabilistic, kiarashinejad2020deep} Created by Kingma \citep{kingma2013auto} it encodes $x$, conditioned on $y$, into a Gaussian distributed random variable $z$. It is a bayesian approach with a proxy loss of Evidence Lower Bound. 
\begin{equation}
    Loss = (x - \hat{x})^2 - \frac{\alpha}{2} \cdot (1 + log\sigma_z + \mu_z^2 - \sigma_z)
\end{equation}
$Z$ (re-parameterized into $\sigma_z, \mu_z$) represents the transformed distribution of hidden state $x$ given $y$. The transformation is learned with trade-off between the reconstruction (decoding back to exactly the same $x$) and distribution ($z$ being normal). cVAE explores the solution space by drawing new examples from $\sigma_z, \mu_z$.  We used the implementation introduced by \citep{ma2019probabilistic} for this approach.

\textbf{Invertible Neural Networks (INN)} \citep{ardizzone2018analyzing}. Invertible Neural Network are based upon the RealNVP \citep{dinh2016density}, and circumvent the one-to-many mapping problem by padding the (assumed) lower-dimensional $y$-space with some random vector $z$, and then learning a bijective transformation between the $x$ and $y\bigotimes z$ (i.e., cross-product) spaces.  There are two ways of training reported in \citep{kruse2019benchmarking}: (i) a supervised L2 reconstruction loss and a Maximum Mean Discrepancy (MMD) \citep{gretton2012kernel}; and (ii) a maximum likelihood estimate (MLE) loss to enforce $z$ to be normally distributed \citep{dinh2016density}. Since the MLE gives a better solution in the literature \citep{ardizzone2018analyzing}, we adopt it here, given by 
\begin{equation}
    Loss = \frac{1}{2} \cdot ( \frac{1}{\sigma^2} \cdot (\hat{y} - y_{gt})^2 + z^2) - log|det J_{x \mapsto[y,z]}|\\
\end{equation}
where $J$ means the Jacobian of mapping from $x$ to $y\bigotimes z$ space and $z$ represents the transformed values of $x$.  Exploration of the inverse solution space is accomplished by sampling $z$ values from a zero-mean Gaussian distribution.  These $z$ values are concatenated to the target $y$ value and passed through the network to obtain an inverse estimate, $x$.  INN requires equal dimensionality of $x$ and $y\bigotimes z$; in cases where this is violated, we follow \cite{ardizzone2018analyzing} and pad wth zeros. 

\textbf{Conditional Invertible Neural Networks (cINN)}. Conditional INNs use a similar network structure as INNs, with a modification that instead of learning the bijective mapping from $x$ to $y \bigotimes z$ space, it learns the bijective relationship between $x$ and $z$ space under condition $y$. The network is trained under MLE loss as well, with the caveat that $y$ does not appear in the loss function due to conditioning.
\begin{equation}
    Loss = \frac{1}{2} z^2 - log|det J_{x \mapsto z}|
\end{equation}
Here $z$ represents the full transformed distribution of $x$ conditioned on $y$. Exploring inverse solution space also requires sampling different $z$ values. We adopted the original author’s implementation in both invertible networks, \citep{kruse2019benchmarking} in order to avoid inadvertent alteration of the comparison condition.

\textbf{Tandem model} \citep{liu2018training,jordan1992forward}.  In this approach a neural network is first trained to approximate $f(x)$ using a standard regression loss (e.g., squared error).  The parameters of $\hat{f}$ are then fixed, and an inverse model $\hat{f}^{-1}(y)$ is pre-prended to $\hat{f}$, and it is trained in an end-to-end manner using backpropagation with the following loss:  
\begin{equation}
    Loss = (\hat{f}( \hat{f}^{-1}(y) ) - y_{gt})^2 + \mathcal{L}_{bnd}
\end{equation}
This loss measures the re-simulation error of each inferred inverse solution, and therefore $\hat{f}^{-1}$ only needs to learn to identify one of the (potentially many) valid inverse solutions to minimize the loss.  As a consequence and, unlike all other inverse models, the Tandem only returns one solution for any given $y$ (i.e., it does not benefit as $T$ grows).  In the appendix we also show that adding the boundary loss, $\mathcal{L}_{bnd}$, during training is highly beneficial for the Tandem model.  

\section{Benchmark Tasks}
We consider four benchmark tasks, which are summarized in Table \ref{tabledatasets}. Inspired by the recent benchmark study \citep{kruse2019benchmarking}, we include two popular existing tasks: ballistics targeting (D1), and robotic arm control (D3).  For these two tasks we use the same experimental designs as \citep{kruse2019benchmarking}, including their simulator (i.e., forward model) parameters, simulator sampling procedures, and their training/testing splits. All details can be found in \citep{kruse2019benchmarking} and our supplement.  The remaining two benchmarks are new, and we describe them next.  
\subsection{A new meta-material benchmark (D4), and a technique for replicating it}
The goal of this task, recently posed in \citep{nadell2019deep}, is to design the radii and heights of four cylinders (i.e., $x \in \mathbb{R}^8$) of a meta-material so that it produces a desired electromagnetic (EM) reflection spectrum ($y \in \mathbb{R}^{300}$), illustrated in Fig. \ref{fig:MM}.  The input and output are (relatively) high-dimensional and non-linear, and $f(x)$ can only be evaluated using slow iterative EM simulators, requiring significant time and expertise.  These challenges are typical of modern (meta-)material design problems, forming a major obstacle to progress.  Substantial recent research has been conducted on similar problems (e.g., \citep{ma2019probabilistic,peurifoy2018nanophotonic,liu2018training,tahersima2018deep,liu2018generative}), making this both a challenging and high-impact benchmark problem.  

Problems like this are not suitable as benchmarks due to the computation time, needed domain expertise, and required use of a simulator. It is also insufficient simply to share data from the simulator, due to the need to draw new samples from $f(x)$ when evaluating inverse models. We overcome this problem by generating a large number of samples from our simulator (approx. 40,000), and then training an ensemble of deep neural networks to approximate the simulator.  This yields a highly accurate simulator (mean-squared-error of 6e-5) that is fast, portable, and easy to use by others.  All of our experiments utilize data sampled from this proxy simulator rather than the original simulator.   We hypothesized that the difficulty of our meta-material problem may be undermined because we use the same class of models (neural networks) for both the proxy-simulator and our inverse models.  We mitigate this risk by providing a much larger set of training data to the simulator model, and using an ensemble of large and varying models for the proxy-simulator.      

\subsection{The 2-dimensional sinusoidal benchmark (D2)}
This benchmark problem consists of a simple 2-dimensional sinusoidal function, of the following form: $y= \sin(3 \pi x_{1}) + \cos(3 \pi x_{2})$. We included this problem because it had both of the following properties: (i) despite its simplicity, we found it is challenging for most of the deep inverse model; (ii) its 2-dimensional input space allowed us to visualize the solutions produced by each inverse model, and study the nature of their errors.  We utilize these properties to gain deeper insights about the inverse models in Section \ref{sec: analysis}.   

\begin{minipage}{0.4\textwidth}
    \begin{table}[H]
        \caption{Benchmarking datasets outline}
        \label{tabledatasets}
        \centering
            \begin{tabular}{cccc}
            \toprule
            ID & Dataset & Dim(x) & Dim(y) \\ 
            \midrule
            D1 &Ballistics  & 4 & 1 \\ 
            D2 &Sine wave & 2  & 1 \\ 
            D3 &Robotic arm  & 4      & 2 \\ 
            D4 &Meta-material  & 8      & 300      \\ 
            \bottomrule
            \end{tabular}
    \end{table}
\end{minipage}
\hfill
\begin{minipage}{0.5\textwidth}
    \begin{figure}[H]
      \includegraphics[width=1\textwidth]{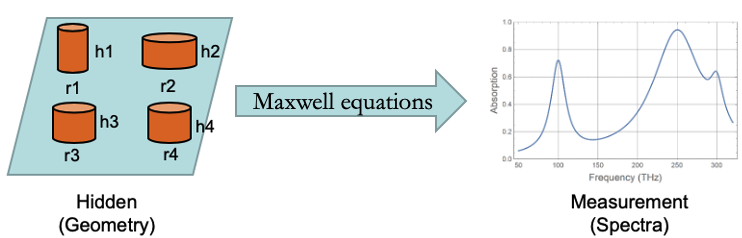}
      \caption{Illustration of the meta-material problem}
      \label{fig:MM}
    \end{figure}
\end{minipage}

\section{Experimental Design and Results} \label{sec: mainresult}
We follow closely the design of the recent benchmark study \citep{kruse2019benchmarking}. For all experimental scenarios that we share with \citep{kruse2019benchmarking}, we followed their design and obtained (with one exception) similar results. This includes results for the cINN, INN and cVAE models; on the Robotic Arm and Ballistics tasks.  In an effort to compare models fairly, we constrained the newly included models -- Tandem and NA -- to have the same number (or less) of trainable parameters.  Furthermore, all models utilized the same training and testing data, batch size, and stopping criteria (for training).  In those cases where model hyperparameters were not available from \citep{kruse2019benchmarking}, we budgeted approximately one day of computation time (on common hardware) to optimize hyperparameters, while again constraining model sizes.  Full implementation details can be found in the supplementary material.  

Once each model was trained, we estimated its error, $r_{T}$ for $T\in\{1,10,20,...,50\}$ using $D=\{x_n,y_n\}_{n=1}^N$ random samples from the simulator. We used the following sample estimator of $r_T$:
\begin{equation}
    \hat{r}_T = \frac{1}{N} \sum_{n=1}^{N} [\min_{z\in Z_{T}} \mathcal{L}(\hat{y}(z),y_{n})]  \label{eq:rtestimate}
\end{equation}
where $Z_{T}$ is a randomly drawn sequence of $z$ values of length $T$.  We use mean-squared error as $\mathcal{L}$, following convention \citep{kruse2019benchmarking,ardizzone2018analyzing}.  A unique set of $z$ values was drawn for each model, based upon the sampling distribution required by that particular model (e.g., Gaussian for cINN).  

\begin{figure}[htb]
  \centering
  \includegraphics[width=12cm]{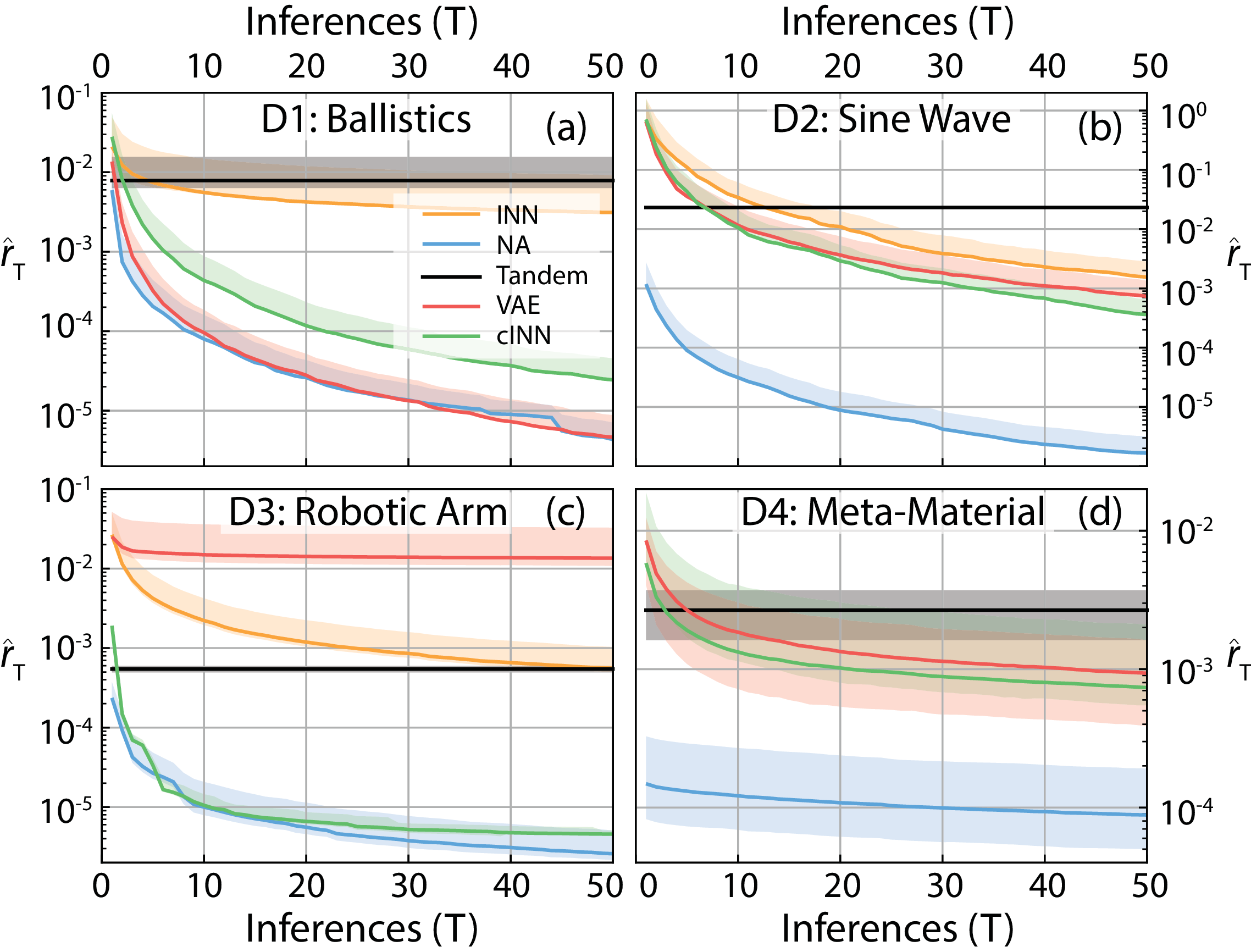}
  \caption{(a-d) Performance on each model for each benchmark task as a function of $T$.}
  \label{fig:Perfcombined}
\end{figure}

The main experimental results are presented in Fig. \ref{fig:Perfcombined}. Measuring $\hat{r}_T$ as a function of $T$ yields a much richer characterization of each model's performance compared to using just $T=1$.  In Fig. \ref{fig:Perfcombined} we see that $\hat{r}_{T}$ falls steadily as $T$ increases, except for the Tandem model with is not stochastic.  Therefore $\hat{r}_{T}$ quantifies the error one can expect for each model depending upon the computational time/hardware permitted for inference available to a user for their application.  Much more interesting is the observation that the performance rank-order of the models also varies with $T$ for all four tasks.  Therefore, the best model for a given task  (in terms of $\hat{r}$) also depends upon the time/hardware permitted for inference. 

\subsection{Which models perform best?} 
\label{sec:bestmodels}
The NA method almost always yields the most accurate solutions, across both tasks and settings of $T$. Especially notable is its large performance advantage on the higher-dimensional meta-material task, suggesting it may be especially effective for similar problems.  However, NA has the drawback of significantly greater computational costs than the other models, due to its use of gradient descent.  The inference time for all model/task combinations is shown in Table \ref{Table:inf}.  We report the time for a single mini-batch of a thousand inferences, which also closely approximates the inference time for a single inference of each model, due to GPU's efficient parallel processing.  Therefore, if one inverse solution can be inferred, then (on standard hardware) many inverse solutions can be obtained in roughly the same amount of time, in which case simulation time becomes the biggest bottleneck (i.e., value of $T$).  

With these computational considerations in mind, as discussed, if enough time is available for at least one inference of NA, then it is the best choice for nearly every task and setting of $T$.  However, for more time-sensitive applications where a single inference from NA is too slow, e.g., many real-time tasks, we are limited to selecting among the other models, which all have (relatively) similar inference time.  Given these similarities, the best choice depends more strongly upon the remaining time available for simulation (i.e., value of $T$).  In this scenario, the Tandem model consistently achieves the best accuracy for time-sensitive applications, where $T$ is small.   If more than a few simulations can be run, then the cINN and the VAE appear to generally achieve the best results: the cVAE performs best on the ballistics task, while the cINN performs best for the robotic arm and sine wave task.

% Conclusion thus far:  Assuming there is enought time for one forward inference, then there is time for many forward inferences, and then simulation immediately becomes the bottleneck.  Therefore, first we ask which applications have enough time for at least one inference of each inverse model?  After that, simulation is the bottleneck and the reader should refer to Fig. 4.    

% Beyond this, things get quite complicated because faster inverse methods will permit more simulations than slower models, but the precise simulation advantage will depend upon (1) the speed difference of the methdos, (2) the time per simulation (and to what extent they can be parallelized, e.g., think CST), and the total allowable processing time (i.e., simulation time plus inverse time).   

\begin{table}[H]
    \caption{Total Inference time (t) in seconds for 1,000 solutions}
    \label{Table:inf}
    \centering
    \begin{tabular}{ccccccc}
        \toprule
        Dataset & NA & Tandem  & cVAE     & INN     & cINN & MDN    \\ 
        \midrule
        D1:Ballistics                  & 1.36  & 0.31 & 0.29 & 0.35 & 0.78 & 0.08\\ 
        D2:Sine wave                   & 1.22  & 0.19 & 0.19 & 0.19 & 0.20 & 0.53\\ 
        D3:Robotic arm                 & 1.12  & 0.19 & 0.31 & 0.21 & 0.23 & 0.62\\ 
        D4:Meta-material               & 46.10  & 0.50 & 0.47 & 0.22  & 0.25 & 0.41\\
        \bottomrule
    \end{tabular}
\end{table}

\begin{table}[H]
    \caption{Estimated Asymptotic Performance of Each Model ($\hat{r}_{T=200}$)}
    \label{Table:asymptoticperformance}
    \centering
    \begin{tabular}{ccccccc}
        \toprule
        Dataset &\textbf{NA} & Tandem  & cVAE     & INN     & cINN & MDN    \\ 
        \midrule
        D1:Ballistics                  & \textbf{2.50e-7}  & 7.84e-3 & 2.80e-7 & 2.20e-3 & 1.18e-6 & 6.6e-6\\ 
        D2:Sine wave                   & \textbf{1.33e-7}  & 
        1.17e-2 & 4.34e-5 & 1.24e-4 & 2.72e-5 & 5.21e-5\\
        D3:Robotic arm                 & \textbf{6.61e-7}  & 5.44e-4 & 1.25e-2 & 2.12e-4 & 8.80e-7 & 1.82e-5 \\ 
        D4:Meta-material               & \textbf{6.67e-5}  & 2.53e-3 & 5.49e-4 & 3.83e-2 & 4.45e-4 & 5.15e-4\\
        \bottomrule
    \end{tabular}
\end{table}

 \begin{wrapfigure}{R}{0.5\textwidth}
  \centering
  \includegraphics[width=0.5\textwidth]{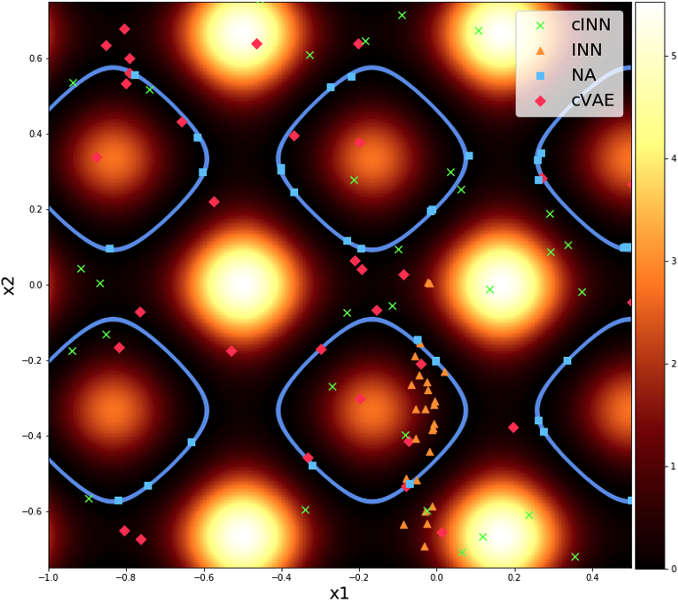}
  \caption{The image axes represent a uniform grid of potential inverse solutions,$(x1,x2)$, for the 2-dimensional sinusoid problem.  The pixel intensity at each $(x1,x2)$ location represents the corresponding simulation error of that solution, if our target measurement is $y=-0.3$. The blue rings represent the optimal solutions.}
  \label{fig:Space Exploration}
\end{wrapfigure}
\subsection{Why does the neural-adjoint perform so well?} \label{sec: analysis}
 Notably in Table \ref{Table:asymptoticperformance}, we see that NA always achieves the lowest asymptotic error as a function of $T$, while the other models asymptote at varying levels.  Why are the other models limited in their accuracy, even as $T\rightarrow \infty$?  One potential explanation is that they do not fully explore $x$-space, and thereby miss some accurate solutions.  Alternatively, perhaps they can find solutions near all of the global optima, but they cannot accurately localize them (e.g., their estimates are noisy). To answer this question, we visualize the 2-dimensional sinusoid task (D2), on which most of the models perform poorly.  Fig. \ref{fig:Space Exploration} presents a random sample of inverse solutions produced by each model, laid on top of a 2-dimensional error map of $x$-space (darker is better). The blue rings indicate the optimal solutions for a target measurement of $y$=-0.3.  We can see clearly that NA finds highly accurate solutions in each of the globally optimal rings.  The cVAE and the cINN seem to find solutions near all of the globally optimal solutions, however they rarely infer perfectly accurate solutions.  Therefore both the cVAE and cINN seem to suffer from noisy solutions, rather than inability to find the solutions.  Finally, the INN seems unable to search the entire space, in addition to suffering from inaccurate solutions.  However, this is a single visualization, representing a single task and a single instance of training the models.  We find the relative performance of all models (except NA) varies substantially across tasks \textit{and} the success of their training (which is somewhat random).  This suggests that each of these models sometimes suffer from limited exploration of $x$-space, and varying accuracy, depending upon the aforementioned factors. These findings are consistent with (e.g., \citep{kruse2019benchmarking}) overall.

 \section{Conclusions} \label{sec: conclusion}
 In this work we presented a large benchmark comparison of five modern deep inverse models, on four benchmark tasks.  We propose a new metric, $r_T$, that evaluates the error of models as a function of the number of inverse solutions they are permitted to propose, denoted $T$.  We find that the performance of inverse models, both in absolute error and in their rank-order, depends strongly on $T$, suggesting that $r_T$ is important to characterize inverse model performance. We also introduce a challenging contemporary inverse problem for meta-material design. Normally, it would be difficult for others to replicate such real-world problems however, we introduce a strategy for creating simple, fast, and sharable \textit{approximate} simulators.  Finally, we propose a method called the Neural-Adjoint, which nearly always achieves the lowest error across all tasks and values of $T$.  Its performance advantage is especially strong for the higher-dimensional meta-material problem, suggesting it is a promising approach to solve such problems.   
 
 %\pagebreak 

\section*{Broader Impact}
 We believe the most proximate impacts of this work will be positive.  In particular, higher-dimensional inverse problems like our meta-material problem present a major obstacle to the development of beneficial technologies across many disciplines e.g., in materials, chemistry, and bio-chemistry.  The Neural-Adjoint method represents a tool to develop much more accurate inverse designs for these complex problems. Furthermore, the ability to replicate inverse studies for complex problems, as we propose, will also accelerate progress, and enable many researchers to study these problems even if they lack sophisticated simulation equipment or expertise. As with many tools, we also acknowledge that these advances can be used to accelerate the development of technologies that are used for negative purposes, which we believe is the most immediate negative outcome of our work.     
 
\acksection
We gratefully acknowledges support from the Department of Energy (DOE) (DE-SC0014372).

\section{Supplementary material}

\subsection{Posterior matching score}
Although the performance over time is the main performance that we want to benchmark, as pointed out by \cite{kruse2019benchmarking} the posterior matching is another metric to measure how good the inverse models are. Below we show the posterior matching score using Maximum Mean Discrepancy (MMD) as a measurement of how close the inferred posterior density is comparing with the ground truth (rejection sampled) distribution. Note that for a real-life problem (D4: meta-material) with higher dimensionality, the rejection sampling becomes intractable. The 3 MMD kernel used was 0.05, 0.2 and 0.9. The code is also available on the repository.

\begin{table}[H]
    \caption{Posterior matching MMD score}
    \label{Table:MMD}
    \centering
    \begin{tabular}{ccccccc}
        \toprule
        Data & NA & TD  & cVAE     & INN     & cINN  & MDN   \\ 
        \midrule
        D1:Ballistics                  & 0.07  & 2.62 & 0.07 & 2.03 & 0.04 & \textbf{0.04}\\ 
        D2:Sine wave                   & 0.04  & 2.84 & 0.03 & 1.07 & 0.03 & \textbf{0.03}\\ 
        D3:Robotic arm                 & 0.06  & 2.70 & 1.62 & 0.11 & 0.04 & \textbf{0.03}\\ 
        \bottomrule
    \end{tabular}
\end{table}

The bold faced models are the best performing ones with respect to posterior matching within one dataset (more digits are compared if ties). We find that Mixture Density Network (MDN) always has the best (lowest) posterior matching MMD score, closely followed by NA and cINN. We hypothesize that MDN wins due to it explicitly optimizes its network weights on posterior matching. Note that the posterior matching does not guarantee good inverse solution on average, as illustrated by the fact that the average re-simulation accuracy of MDN is actually far from that of NA.

\subsection{Neural Adjoint (NA) ablation studies}

In this subsection we present an ablation study for (i) the boundary loss, $\mathcal{L}_{bnd}$, and (ii) the design of the distribution from which we draw initial z, $\Omega$ in the Neural-Adjoint method.  Our goal is to show experimental evidence that these additions to NA generally improve its performance.  Note that the details of these two methods are provided in the main manuscript. In our ablation experiments we evaluate the performance of NA as we remove or include each of these steps, as shown in Table \ref{Table:naablationtable} (left-most three columns).  In those cases where we do not design $\Omega$, we set $\Omega$ to a uniform sampling distribution.  Aside from the specific experimental variables listed in Table \ref{Table:naablationtable}, these experiments all follow the experimental design outlined in the main paper. 

\begin{table}[H]
    \caption{Ablation Study Experimental Design}
    \label{Table:naablationtable}
    \centering
    \begin{tabular}{ccccc}
        \toprule
        Label & Add $\Omega$?  & Add $\mathcal{L}_{bnd}$? & $\hat{r}_{200}$ for D1: ballistics & $\hat{r}_{200}$ for D3: robotic arm \\ 
        \midrule
          E1 & No  & No & 4.78e0 & 1.53e-6 \\ 
          E2 & No  & Yes & 3.54e-6 & 8.87e-7\\ 
          E3 & Yes  & No & 1.39e0 & 5.70e-6\\ 
          E4 & Yes  & Yes & \textbf{2.54e-7} & \textbf{6.61e-7} \\ 
        \bottomrule
    \end{tabular}
\end{table}

\begin{figure}[H]
  \centering
  \includegraphics[width=13.5cm]{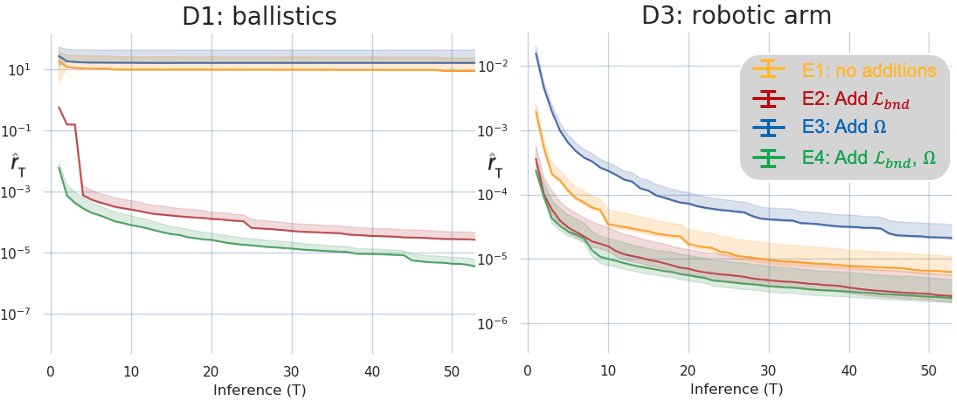}
  \caption{Effect for with or without Boundary loss and prior initialization}
  \label{fig:NAablation}
\end{figure}

We conduct these four experiments on two tasks: the ballistics task (D1) and the robotic arm control task (D2) and Fig. \ref{fig:NAablation} presents the results of these experiments in terms of $r_{T}$ as $T$ varies from one to fifty.  The asymptotic performance of each model is estimated by $r_{T=200}$, and is presented in Table \ref{Table:naablationtable} (right-most two columns).  The results indicate that adding both steps to NA (i.e., transitioning from E1 to E4) results in substantial performance improvements for both tasks considered.  For the ballistics task (D1) there is a reduction in error by several orders of magnitude, and for the robotic arm task (D2) there is a reduction in error of 1-1.5 orders of magnitude.  The most important step appears to be the boundary loss which, by itself, results in substantial performance improvements.  Adding the $\Omega$ design to the boundary loss results in a smaller, but consistent performance improvement.  Interestingly we find that adding the $\Omega$ design without $\mathcal{L}_{bnd}$ is detrimental on both tasks.  

\subsection{Neural Adjoint (NA): Visualization of why boundary loss helps}
As illustrated in the main paper, the neural adjoint has implementation caveat where constraining the boundary of the solution search phase is crucial to the performance of the inverse problem solving. Here to visualize how and why boundary constraint plays such an important role, we would use a simple toy example of fitting a 1d sine wave (x from -$\pi$ to $\pi$) using a small neural network.

\begin{figure}[ht]
  \begin{subfigure}[t]{0.5\textwidth}
      \includegraphics[height=6cm]{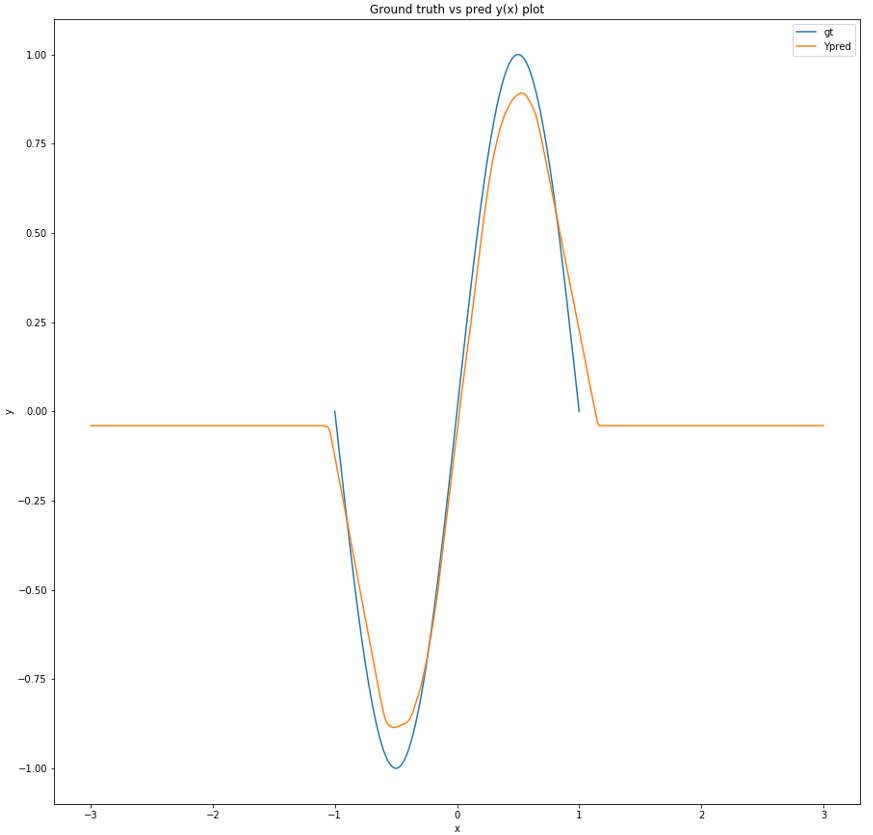}
      \caption{Approximating a simple 1d sine function (x in unit of $\pi$)}
      \label{fig:simple sine}
  \end{subfigure}
  \begin{subfigure}[t]{0.5\textwidth}
      \includegraphics[height=6cm]{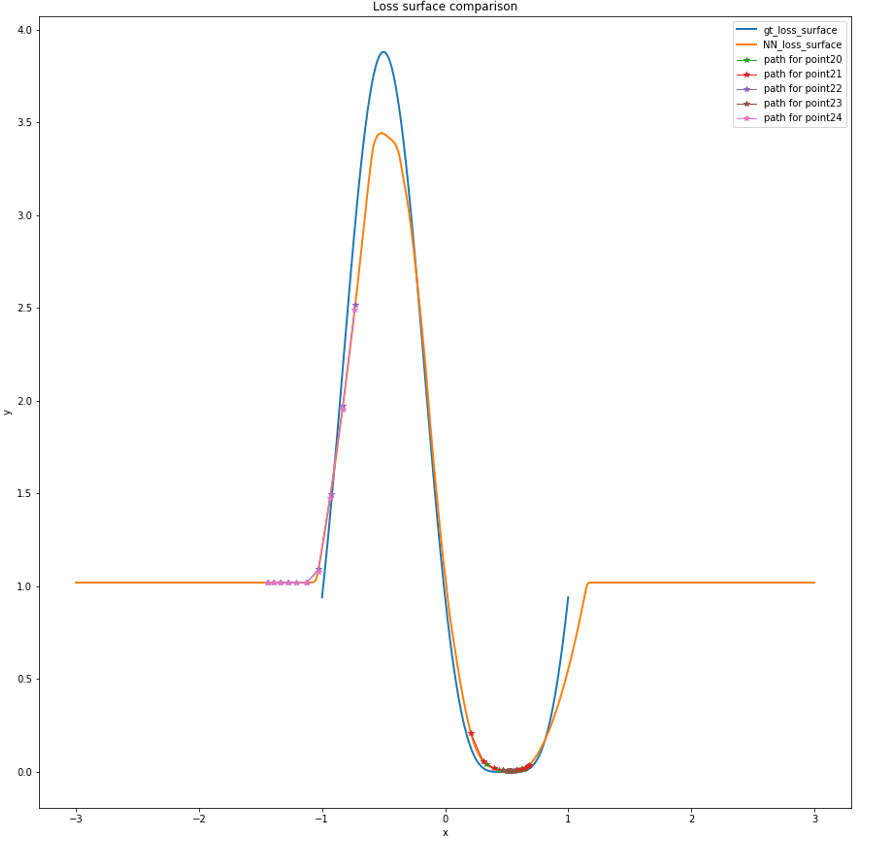}
      \caption{Error surface in searching for solution y=0.6}
      \label{fig:errorsurface1}
  \end{subfigure}
  \caption{First trail of toy data on 1d sine wave dataset}
\end{figure}

As shown in fig \ref{fig:simple sine} it fits pretty well in the range $[-\pi, \pi]$ and seem to choose an arbitrary small value for out of range domain $(-\infty, -\pi] \cup [\pi, \infty)$. Then we do a neural adjoint method searching for the x value for given y value 0.6 and visualize the error surface of the search in fig \ref{fig:errorsurface1}. The figure shows how initial guesses of points would be guided by gradient and move towards lower points on this loss surface graph. As several points in the right half found the global minimum, the point initialized to the left struggles to find a lower result. It seems that the left point's failure to find the global minimum would not cause trouble as we can choose those points who have a lower error in a parallel run to avoid points stuck in local minimum like this. 

\begin{figure}[ht]
  \begin{subfigure}[t]{0.5\textwidth}
      \includegraphics[height=6cm]{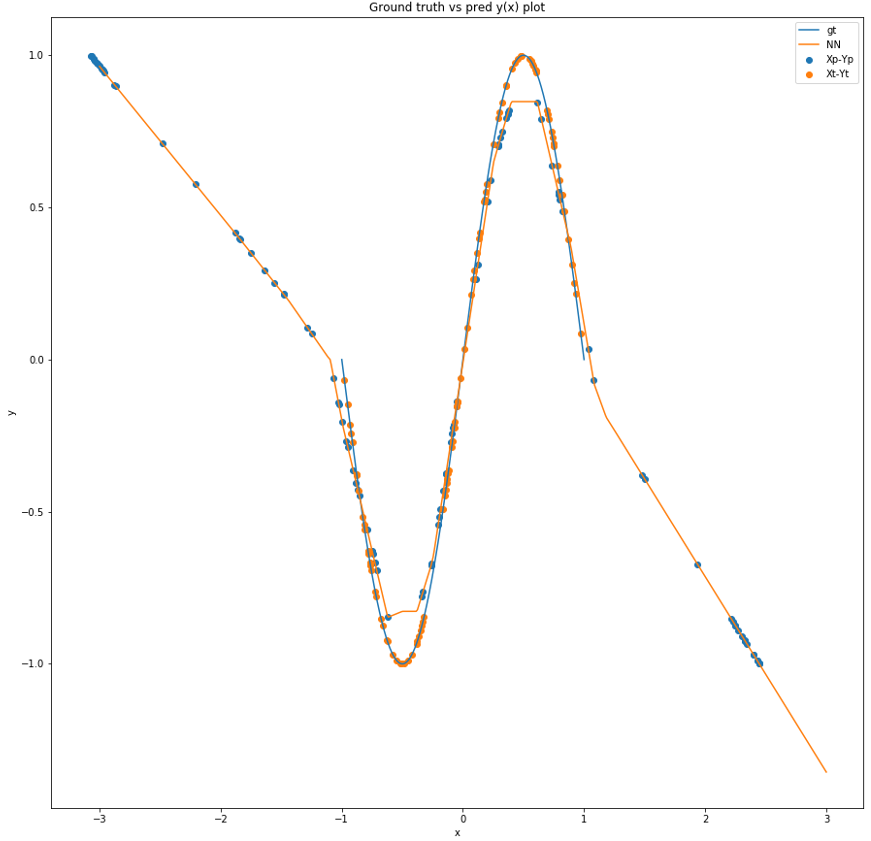}
      \caption{Prediction chart of 1d sine wave experiment run 2}
      \label{fig:1dsinerun2}
  \end{subfigure}
  \begin{subfigure}[t]{0.5\textwidth}
      \includegraphics[height=6cm]{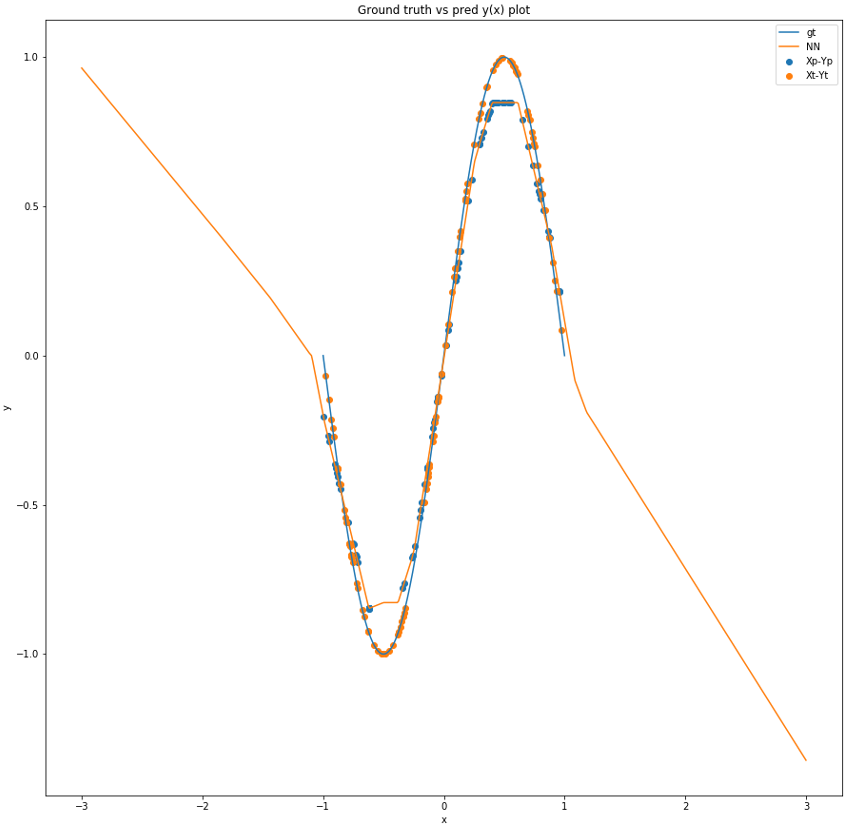}
      \caption{After adding boundary error during inference in 1d sine wave experiment run 2}
      \label{fig:afterboundary}
  \end{subfigure}
  \caption{Second trail of toy data on 1d sine wave dataset}
\end{figure}

However, if we run the whole experiment one more time, it is a different story. As shown in fig \ref{fig:1dsinerun2}, the orange line still symbolizes what the neural network learns and the blue line is the ground truth. The blue points are various inverse solution found corresponding to their ground truth point in orange. All those solutions found are deemed global minimum with 1e-1 of error seen by the network. this time the network "decided" not to learn close-to-0 values for out-of-range $(-\infty, -\pi] \cup [\pi, \infty)$ domain. Instead, it learns an affine function at both tails and that unpredictable behavior caused the trouble as some of the solutions are actually out-of-range, thus meaningless and cause a high re-simulation error (0.7).

As said in the main paper, we solved this with an extra boundary loss which bound the inference solution exploration within the defined range. With the added loss, illustrated in fig \ref{fig:afterboundary}, the out-of-range inverse solutions disappear and the re-simulation error converges to the forward approximation error (4e-3).

\subsection{Tandem model ablation study} \label{TD:boundary}
\begin{figure}[H]
  \centering
  \includegraphics[width=13.5cm]{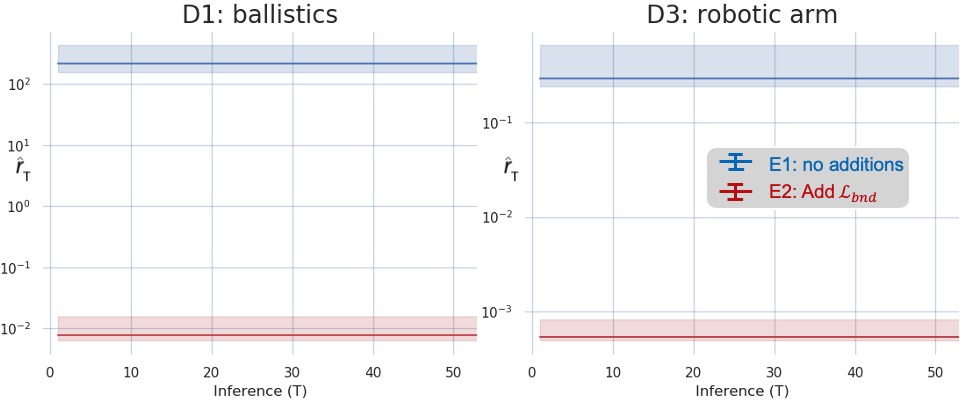}
  \caption{Effect for with or without Boundary loss for Tandem model}
  \label{fig:TDablation}
\end{figure}

Due to the success of the boundary loss, $\mathcal{L}_{bnd}$, within the Neural Adjoint approach, we also considered whether it may be beneficial for the Tandem model as well.  To test this, we evaluate the performance of the Tandem model with, and without, the inclusion of $\mathcal{L}_{bnd}$.   We conduct these two experiments on two tasks: the ballistics task (D1) and the robotic arm control task (D2).  Aside from excluding the boundary loss, we use the same experimental design used in the main paper.  The results of the experiments are presented in Fig \ref{fig:TDablation}.  the results indicate that inclusion of $\mathcal{L}_{bnd}$ has substantial and consistent benefits, reducing error by at least 2.5 orders of magnitude on both tasks. 

\subsection{Benchmark Deep Inverse Models: Additional Details}
In this subsection we provide additional technical details for each of the existing benchmark inverse models employed in our main paper.  
\iffalse
\subsubsection{Tandem Model (TD)}
The tandem model training takes 2 phases. The first training phase works as the Neural adjoint method where an accurate forward mapping is learned. The second phase of training fixes the weight and hence make use of the first phase training result by producing much more stable loss feedback toward the backward model. In the inference phase, the tandem model would use only the backward model to infer and it would give deterministic inverse solution x for each target input y.
\begin{figure}[H]
  \centering
  \includegraphics[width=13cm]{Arch_v3_figs/TD.png}
  \caption{Architecture of Tandem method (white blocks and arrows indicate not active in current phase).}
  \label{fig:TD}
\end{figure}
To train a tandem models, L2 MSE loss is used with an additional boundary loss mentioned in equation \ref{TD:boundary}. The network can be composed of fully connected layer in simple datasets (D1-D3) or have convolutional layers (D4) with the upconvolution layer in the opposite side.

\begin{equation}
    Loss = (\hat{f}(\hat{x}) - y_{gt})^2 + \mathcal{L}_{bdy}
\end{equation}
\fi

\subsubsection{Conditional Variational Auto-Encoder (cVAE)}
The conditional variational auto-encoder adopts the evidence lower bound as it encodes the x into gaussian distributed random variable z conditioned on y. During training phase it also make use of the L2 MSE loss to ensure a good reconstruction of the original input x. During inference phase, inverse solution x is decoded from random samples are drawn from z space conditioned on y.
\begin{figure}[H]
  \centering
  \includegraphics[width=13cm]{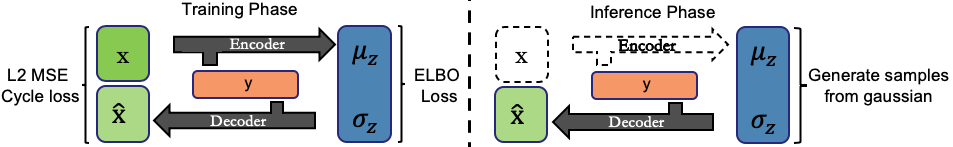}
  \caption{Architecture of conditional Variational Auto-Encoder method}
  \label{fig:cVAE}
\end{figure}
With the evidence lower bound loss defined in equation \ref{ELBO}, it trades-off between the reconstruction of the original signal and the shape of the distribution of the latent variable z. Upon minimization of the loss the network is supposed to fully represent the joint distribution using a normally distributed latent space.
\begin{equation}
    Loss = (x - \hat{x})^2 - \frac{\alpha}{2} \cdot (1 + log\sigma_z + \mu_z^2 - \sigma_z) \label{ELBO}
\end{equation}

\subsubsection{Invertible neural network (INN)}
The invertible neural network is specially designed to have hard invertibility (full reconstruction). During training, it uses Maximum likelihood loss to map the bigger x space into y and z space, where z is sampled from a normal distribution. During the inference phase, a randomly drawn normal distributed z would join y to be inverted back to the inverse solution x.

\begin{figure}[H]
  \centering
  \includegraphics[width=13cm]{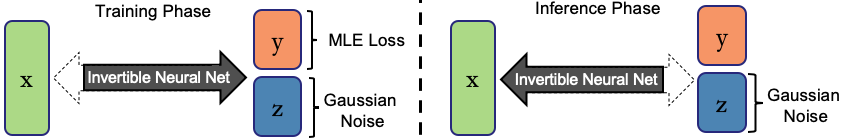}
  \caption{Architecture of Invertible Neural Network method}
  \label{fig:INN}
\end{figure}

By taking the assumption that y is normally distributed around its ground truth value, the network can be trained using simple maximum likelihood loss defined in equation \ref{INN:MLE}. To ensure the invertibility the Jacobian of the transformation is also added to the loss, encouraging full invertibility upon convergence. The variance $\sigma$ is set to be small to encourage accuracy and is chosen based on cross validation.

\begin{equation}
    Loss = \frac{1}{2} \cdot ( \frac{1}{\sigma^2} \cdot (\hat{y} - y_{gt})^2 + z^2) - log|det J_{x \mapsto[y,z]}|\\ \label{INN:MLE}
\end{equation}

\subsubsection{Conditional invertible neural network (cINN)}
The conditional invertible neural network uses a similar structure as an invertible neural network. Instead of mapping x to yz space, by conditioning on y, it approximates the full mapping between x and a normally distributed random variable z using maximum likelihood as well. During inference, a normally distributed random variable would be drawn to get inverse solution x conditioned on y.
\begin{figure}[H]
  \centering
  \includegraphics[width=13cm]{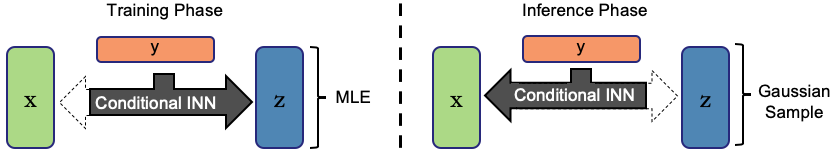}
  \caption{Architecture of conditional INN method}
  \label{fig:cINN}
\end{figure}
Very much like the Invertible neural network above in equation \ref{INN:MLE}, the conditional version use equation \ref{CINN:Loss} as loss function since it already have y information given.
\begin{equation}
    Loss = \frac{1}{2} z^2 - log|det J_{x \mapsto z}| \label{CINN:Loss}
\end{equation}

\subsubsection{Mixture density network (MDN)}
Proposed by \cite{bishop1994mixture}, Mixture density network provides a simple model for one-to-many relationships by assuming a gaussian mixture for the posterior density where the mean and maximum of the gaussians are determined by the input y. The number of gaussian mixtures is part of the hyperparameters of the network and is tuned by cross validations.

\begin{figure}[H]
  \centering
  \includegraphics[width=13cm]{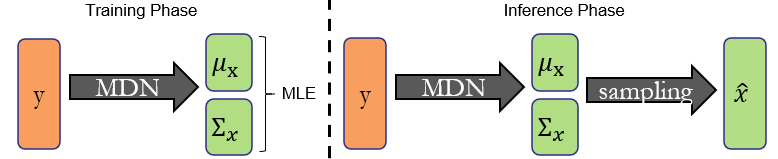}
  \caption{Architecture of MDN method}
  \label{fig:cINN}
\end{figure}
It is trained using a maximum likelihood method and during inference a guassian sampling is done to retrieve the estimate of x.
\begin{equation}
    Loss = - \log(  \sum_i p_i *|\Sigma_i^{-1}|^{\dfrac{1}{2}} * \exp(-\dfrac{1}{2} (\mu_i - x)^{T} \Sigma_i^{-1} (\mu_i - x)))
\end{equation}

\subsection{Benchmark tasks: additional details}
In this subsection we provide additional technical details for each of the benchmark tasks included in the main paper, except for the 2-dimensional sinusoid task, due to its simplicity.  Two of our tasks are adopted directly from the recent deep inverse model benchmark study \cite{kruse2019benchmarking}: the ballistics task and robotic arm control task.  The full details of these benchmarks can be found in \cite{kruse2019benchmarking} but we reproduce them here for completeness.  

\subsubsection{The ballistics task}
A physically motivated dataset as a ball is thrown from position $(x_1, x_2)$ with angle $x_3$ and velocity $x_4$ and land on ground at location $y$. There is no closed form mapping as getting y from given x requires the solve the below equation. Parameter priors are as follows: $x_1 \sim \mathcal{N}(0, \frac{1}{4}), x_2 \sim \mathcal{N}(\frac{3}{2}, \frac{1}{4}), x_3 \sim \mathcal{U}(9 \degree, 72\degree)$ and $x_4 \sim$ Poisson(15).
\begin{align*}
    T_1(t) &= x_1 - \frac{v_1m}{k} \cdot (e^{-\frac{kt}{m}} - 1) \\
    T_2(t) &= x_2 - \frac{m}{k^2} \cdot ((gm + v_2k) \cdot (e^{-\frac{kt}{m}} - 1) + gtk)\\
    y &= T_1(t^*)\ s.t.\ T_2(t^*) = 0\\
\end{align*}

\subsubsection{The robotic arm control task}
Raised by \cite{ardizzone2018analyzing}, it is a simple geometrical problem asking for the starting height $x_1$ and three joint angles $x_{2,3,4}$ given the robotic arm's final position $[y_1, y_2]$. The closed form relationship is as follows with $l_{1,2} = 0.5, l_3 = 1, \textbf{x} \sim \mathcal{N}(0,\boldmath{\sigma^2})$ where $\boldmath{\sigma^2} = [\frac{1}{16}, \frac{1}{4}, \frac{1}{4}, \frac{1}{4}]$.
\begin{align*}
    y_1 &= l_1 sin(x_2) + l_2 sin(x_3 - x_2) + l_3sin(x_4-x_3-x_2) + x_1\\
    y_2 &= l_1 cos(x_2) + l_2cos(x_3-x_2) + l_3cos(x_4-x_3-x_2)\\
\end{align*}

\subsubsection{Meta-material task and approximated simulator}
Our meta-material (MM) task is follows the recent work in \cite{nadell2019deep}, where the goal was to choose a set of geometric parameters for a MM design so that the resulting MM exhibits some desired electromagnetic properties.  In our context, MMs consist of a surface (e.g., a semiconductor wafer) with small repeating geometric structures (e.g., cylinders, crosses) placed on its surface. The characteristics of these structures (e.g., shape, size, thickness) influence the electromagnetic properties of the resulting MM.  Our particular MM is composed of a repeating "super-cell" of four cylinders, each with two parameters that we can control: a height and a radius. The electromagnetic property we wish to control is called the reflection spectrum, which is a 300-dimensional vector of values between zero and one.  Each value of the reflection spectrum indicates the proportion of signal energy (at each frequency) from an incident electromagnetic ray that would be reflected from the MM surface. In this work our reflection spectrum consists of 300 uniformly-spaced measurements across the frequency range from 0.8 to 1.5 THz, following \cite{nadell2019deep}.

While there is currently no known closed-form mathematical expression for the forward model of this system, $f$, we can still evaluate $f$ for a given $x$ using electromagnetic simulation software. Following \cite{nadell2019deep}, we use the CST Studio simulation software for this purpose.  Although CST is a powerful tool that enables us to study this problem, there are two significant difficulties with using CST (or similar simulators) when studying inverse problems.   These two difficulties impede our study, and also prevent others from replicating our experiments.   First, setting up the simulations requires substantial domain expertise that will not be easily accessible to most researchers. The second problem is that evaluating $f$ is relatively slow.  Like many simulators, CST evaluates the forward model by iteratively solving a differential equation, in our case Maxwell's equation, which is a relatively slow process. For our particular application, CST can produce (approximately) 1000 simulations per day on a single CPU core, which is about 1 simulation every 1.5 minutes.  To carry out our experiment we have 1,000 test points ($y$ values), and we extract 100 proposed solutions ($x$ values) from 4 models for each test point, resulting in 400 days of simulations!

To overcome this problem, following \cite{nadell2019deep}, we trained a neural network to closely approximate the CST simulator.  Although this approach still required substantial computation time and expertise, overall it required far fewer simulations to generate the data needed to train our "neural simulator" than our inverse modeling benchmarks.  Furthermore, we only needed to perform this procedure once, after which we can conduct our experiments much faster using the neural simulator.  In addition to being fast, the neural simulator requires little expertise to use by other researchers, making it both fast and easy to use.  Therefore the neural simulator approach enables other researchers to easily study this previously inaccessible modern inverse problem. Many important modern inverse problems in engineering and research rely on simulators with the same limitations, preventing widespread study of many problems and slowing scientific progress.  We propose this approach as a general strategy to make these complex modern inverse problems accessible to the broader scientific community.  
\begin{figure}[h]
  \centering
  \includegraphics[width=\textwidth]{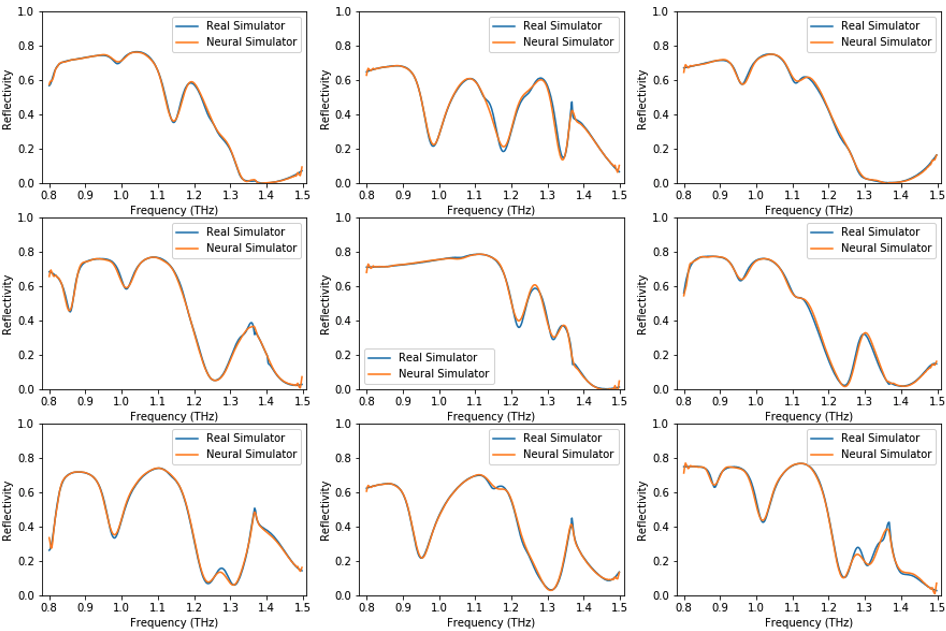}
  \caption{Random Samples from Neural Simulator spectra}
  \label{fig:Neuralsimulator}
\end{figure}

For our particular neural simulator, we randomly sampled 40,000 geometry values from a uniform distribution as discussed in \cite{nadell2019deep}.  We then used CST to generate corresponding reflection spectra for each of these geometry.  Due to inherent symmetry in the parameterization of the meta-material geometry, one can identify several values of $x$ that all correspond to the exact same physical layout of the MM (not discussed \cite{nadell2019deep}).  Leveraging this symmetry we were able to expand the total dataset to 160,000 samples without running additional simulations.  We split the resulting dataset into two subsets: $80\%$ for training and $20\%$ for testing.  Our neural simulator is composed of an ensemble of deep neural network regression models with varying architectures.  After training, our neural simulator achieves a mean-squared error of 6e-5 on the test set.  Some randomly drawn test spectra are presented in Fig. \ref{fig:Neuralsimulator}, along with the predicted spectra from the proxy simulator, providing a qualitative illustration its accuracy. The proxy simulator is extremely fast, capable of producing thousands of forward model evaluations per second.  We subsequently used the proxy simulator to generate all of the data in our experiments.  We release the proxy simulator with this publication.

\iffalse
    \subsection{Different sources of one-to-manyness}
    The crux of the ill-posedness of a lot of inverse problems comes from the one-to-many mapping from inverse direction. Therefore, a better understanding of the origin of that one-to-manyness is of great interest. Here we identify 2 distinct (but not exclusive) properties of problem that would result in a one-to-many mapping.
    \subsubsection{Periodicity}
    Back to our simple example of 1d sine wave problem: $y=\sin (x)$. Whenever the range of x is larger than $\pi$, there are definitely one-to-many in the inverse mapping where two different $x_i$ and $x_i + 2\pi$ maps to the same y value. 
    
    \subsubsection{Dimension mismatch}
    The dimension mismatch between input and output often leads to one-to-many mapping. Imagine you want to infer the previous x value from a reduced dimension of $x^'$ or $y$ by principle component analysis, since you have less dimension to store information, the loss in information generally cause the one-to-many mapping as any change in that direction of information loss would eventually give the same y. This is one of the implicit assumptions of the models that make use of the invertible structure introduced by \cite{dinh2016density}.
\fi

\subsection{Experimental design: additional details}
As discussed in the main body of the paper, our experimental design is based closely upon the recent benchmark study in \cite{kruse2019benchmarking}.  In particular, we shared two benchmark tasks (Ballistics and Robotic Arm) and three inverse models (cVAE, INN, and cINN) with the study in \cite{kruse2019benchmarking}.  For these particular scenarios we followed their task design, and deep model designs (e.g., architectures and hyperparameters) in all cases in which it was specified.  We were able to obtain largely similar results for these common scenarios.  

Table \ref{table:experimentaldesign} presents several additional model training details that we used. These training details remained fixed across all tasks and all models in our experiments. We found that these settings allowed all models to converge before training stopped. With these settings we also were able to find similar error rates to those of \cite{kruse2019benchmarking} on those scenarios that were shared between this work and their work.  

For each model and dataset combination, we allocated one day of GPU processing time to optimize the model.   For those model/task combinations from Kruse, we did not optimize all model parameters that were already specified.  We optimized the remaining parameters (e.g., regularization, multi-task loss weights) but we found these had little impact on our results.  For the remaining models/tasks that were not specified in \cite{kruse2019benchmarking} we also considered optimizing model architectures, while remaining within the same overal processing budget.   All specifications for our models, and the code used to train them, will be published with our paper.  

\begin{table}[h]
    \centering
    \caption{Table for experimental setups}
    \label{table:experimentaldesign}
    \begin{tabular}{cc}
        \toprule
        Parameters          & value\\
        \midrule
        Training Epoch         & 500                                                           \\ 
        Batch size             & 1024                                                          \\ 
        Optimizer              & Adam                                                          \\ 
        Learning rate          & 1e-3                                                          \\ 
        Learning rate schedule & half when plateau \\ 
        Optimization time      & 1 GPU*day                                                     \\ 
        GPU                    & NVIDIA 1080 Ti                                                \\ 
        \bottomrule
    \end{tabular}
\end{table}

\subsection{Additional miscellaneous results}
These additional details were not specifically cited or referenced in the main body of the paper, but we provide them here to supplement the paper.  

\subsubsection{Average re-simulation error (T=1) Performance}
Due to limited space, we did not include the numerical $\hat{r}_T=1$ in the main paper. The $\hat{r}_T=1$ is illustrated in table \ref{Table:rt0performance}.

\begin{table}[H]
    \caption{Estimated Average Performance of Each Model $\hat{r}_T=1$}
    \label{Table:rt0performance}
    \centering
    \begin{tabular}{ccccccc}
        \toprule
        Dataset &\textbf{NA} & Tandem  & cVAE     & INN     & cINN & MDN    \\ 
        \midrule
        D1:Ballistics                  & \textbf{5.00e-3}  & 7.84e-3 & 1.35e-2 & 2.09e-2 & 2.78e-2 & 9.88e-2\\ 
        D2:Sine wave                   & \textbf{1.18e-3}  & 2.31e-2 & 7.56e-1 & 6.70e-1 & 6.45e-1 & 4.46e-1\\
        D3:Robotic arm                 & \textbf{2.00e-4}  & 7.00e-4 & 2.51e-2 & 2.66e-2 & 2.01e-3  & 4.81e-3\\ 
        D4:Meta-material               & \textbf{2.50e-4}  & 2.53e-3 & 8.60e-3 & 3.89e-2 & 5.70e-3 & 4.60e-3\\
        \bottomrule
    \end{tabular}
\end{table}

\subsubsection{Model size}
To cross-validate our result with Kruse \cite{kruse2019benchmarking} , we used models of similar size in the two benchmark problems that we share. For other models and datasets, we decided the size of the model by doing a hyper-parameter swiping and chose best performing model complexity. As shown in Table \ref{Table:numberoftrainable} and Table \ref{Table:modelsize}, invertible structures tends to have a larger number of parameters to model complicated invertible relationships while the NA method, as it only needs to model the one-to-one relationship, requires substantially smaller network structures. 

\begin{table}[H]
    \caption{Model Size in number of free parameters (Millions) }
    \label{Table:numberoftrainable}
    \centering
    \begin{tabular}{ccccccc}
        \toprule
        Dataset &\textbf{NA} & Tandem  & cVAE     & INN     & cINN & MDN    \\ 
        \midrule
        D1:Ballistics                  & \textbf{0.5}  & 0.5 & 3.0 & 3.2 & 3.2 & 3.0 \\ 
        D2:Sine wave                   & \textbf{0.7}  & 1.5 & 2.5 & 2.1 & 5.3 & 4.0 \\
        D3:Robotic arm                 & 0.8  & \textbf{0.3} & 3.0 & 2.6 & 3.2 & 1.5 \\ 
        D4:Meta-material               & \textbf{3.1}  & 3.4 & 19.0 & 7.2 & 11.8 & 6.0 \\
        \bottomrule
    \end{tabular}
\end{table}

\begin{table}[H]
    \caption{Saved Model Size (Mb)}
    \label{Table:modelsize}
    \centering
    \begin{tabular}{ccccccc}
        \toprule
        Dataset &\textbf{NA} & Tandem  & cVAE     & INN     & cINN & MDN    \\ 
        \midrule
        D1:Ballistics                  & \textbf{2}  & 3 & 14 & 15 & 15 & 14\\ 
        D2:Sine wave                   & \textbf{3}  & 7 & 12 & 9.7 & 24& 19 \\
        D3:Robotic arm                 & 3  & \textbf{2} & 7 & 12 & 15 & 7 \\ 
        D4:Meta-material               & \textbf{15}  & 16 & 88 & 54 & 56 & 31\\
        \bottomrule
    \end{tabular}
\end{table}

\subsubsection{Model training time comparison}
The training time for each algorithms are reported under single NVIDIA 1080 GTX GPU. From Table \ref{Table:training time} one can see clear trend that NA method method tends to need less training time, which is expected due to their smaller model size. 
\begin{table}[H]
    \caption{Training Time (s)}
    \label{Table:training time}
    \centering
    \begin{tabular}{ccccccc}
        \toprule
        Dataset &\textbf{NA} & Tandem  & cVAE     & INN     & cINN & MDN    \\ 
        \midrule
        D1:Ballistics                  & \textbf{86}  & 168 & 155 & 345 & 987 & 187\\ 
        D2:Sine wave                   & \textbf{82}  & 135 & 110 & 191 & 291 & 241\\
        D3:Robotic arm                 & \textbf{70}  & 127 & 227 & 932 & 663 & 120\\ 
        D4:Meta-material               & 224  & 256 & 540 & 335 & 882 & \textbf{122}\\
        \bottomrule
    \end{tabular}
\end{table}

\iffalse
    \subsection{Distribution of error}
    Besides the mean squared error, the distribution of error of each model on all dataset are also reported in this supplementary material.
    \begin{figure}[H]
      \centering
      \includegraphics[width=9cm]{Perf_v2_figs/ballisticsBox_plot.png}
      \caption{Box plot of error for ballistics dataset}
      \label{fig:Box ballistics}
    \end{figure}
    
    \begin{figure}[H]
      \centering
      \includegraphics[width=9cm]{Perf_v2_figs/robotic_armBox_plot.png}
      \caption{Box plot of error for robotic arm dataset}
      \label{fig:robotic arm}
    \end{figure}

    \begin{figure}[H]
      \centering
      \includegraphics[width=9cm]{Perf_v2_figs/meta_materialBox_plot.png}
      \caption{Box plot of error for meta-material dataset}
      \label{fig:Box meta material}
    \end{figure}
\fi

\bibliographystyle{ieeetr}
\bibliography{reference}
\end{document}